# Deep Learning-Based Computer Vision Models for Early Cancer Detection Using Multimodal Medical Imaging and Radiogenomic Integration Frameworks

Emmanuella Avwerosuoghene
Oghenekaro
Department of Computer
Science,
University of Illinois at
Springfield,
USA

**Abstract**: Early cancer detection remains one of the most critical challenges in modern healthcare, where delayed diagnosis significantly reduces survival outcomes. Recent advancements in artificial intelligence, particularly deep learning, have enabled transformative progress in medical imaging analysis. Deep learning-based computer vision models, such as convolutional neural networks (CNNs), transformers, and hybrid attention architectures, can automatically extract complex spatial, morphological, and temporal patterns from multimodal imaging data including MRI, CT, PET, mammography, histopathology, and ultrasound. These models surpass traditional radiological assessment by identifying subtle tissue abnormalities and tumor microenvironment variations invisible to the human eye. At a broader scale, the integration of multimodal imaging with radiogenomics linking quantitative imaging features with genomics, transcriptomics, and epigenetic biomarkers has introduced a new paradigm for personalized oncology. This radiogenomic fusion allows the prediction of tumor genotype, immune response, molecular subtypes, and treatment resistance without invasive biopsies. By incorporating multi-omics data and imaging biomarkers into deep learning frameworks, clinicians can generate patient-specific risk scores, detect early tumor onset, and forecast disease progression with high sensitivity and specificity. Narrowing down, this research explores deep learning-based computer vision models that fuse imaging and genetic data using architectures such as multi-stream CNNs, graph neural networks, and transformer-based radiogenomic encoders. These frameworks leverage feature-level and decision-level fusion to correlate radiomic phenotypes with genomic signatures, enabling early diagnosis of cancers such as glioblastoma, breast, lung, colorectal, and prostate cancers. Additionally, challenges including data heterogeneity, interpretability, limited annotated datasets, and ethical concerns surrounding genomic privacy are addressed. The study emphasizes the need for standardized imaging protocols, federated learning systems, and clinically validated AI pipelines to ensure accurate, reproducible, and globally deployable cancer detection systems.

## 1. INTRODUCTION
### 1.1 Global Cancer Burden and Mortality Statistics

Cancer remains one of the leading causes of death worldwide, responsible for nearly 10 million deaths annually, representing one in every six deaths [1]. The World Health Organization reported approximately 19.3 million new cancer cases in 2020, with projections estimating a rise to 28.4 million by 2040 due to aging populations, lifestyle transitions, and environmental exposures [2]. Low- and middle-income countries account for 70 percent of cancer deaths, primarily due to limited access to screening, diagnosis, treatment, and palliative care [3]. Sub-Saharan Africa faces a particularly alarming surge in cancers such as breast, cervical, liver, and prostate, often diagnosed at advanced stages when prognosis is poor [4]. Childhood cancers are also increasing, yet survival remains far lower compared to high-income regions. In addition to human suffering, cancer imposes substantial economic consequences. Global productivity losses and healthcare costs attributable to cancer exceed 1.16 trillion USD each year [5]. Disparities in healthcare infrastructure, public awareness, and oncology workforce availability further exacerbate mortality rates in developing regions [6]. Urbanization, tobacco use, dietary shifts, obesity, and pollution contribute significantly to rising incidence trends. Without rapid advancement in early detection, precision therapy, and multidisciplinary healthcare systems, mortality rates will continue to escalate. Cancer is therefore not just a medical challenge, but a socioeconomic and developmental issue that requires global collaboration, policy commitment, and innovative technologies for sustainable control [7]. Addressing this evolving burden demands new diagnostic methods, equitable resource distribution, and adoption of artificial intelligence-driven tools each year globally.

**1.2 Importance of Early Detection for Improved Prognosis**

Early detection of cancer dramatically improves survival outcomes, reduces treatment costs, and enhances patient quality of life by enabling timely intervention before metastasis occurs [8]. When cancers such as breast, colorectal, and cervical are diagnosed at localized stages, survival rates can exceed 90 percent, compared to less than 30 percent for advanced-stage presentations [4]. Early stage tumors are often smaller, less invasive, and more responsive to surgery, radiotherapy, or targeted therapies. Moreover, early diagnosis allows clinicians to preserve organ function, minimize chemotherapy toxicity, and improve psychosocial outcomes for patients and families [6]. Screening programs, including mammography, Pap smears, colonoscopy, and low-dose CT for lung cancer, have contributed significantly to reduced mortality in high-income nations [2]. However, these programs are less accessible in developing countries, where late presentation is common due to poor awareness, cultural stigma, inadequate screening infrastructure, and limited pathology services. Detecting cancer early also supports better treatment planning through staging, genomic profiling, and identification of precision therapy targets [1]. It further lowers national healthcare expenditures by reducing the need for complex surgeries, extended hospitalization, or palliative care. Despite these benefits, many cancers remain asymptomatic in early stages, making detection highly dependent on advanced imaging and molecular biomarkers [5]. Consequently, there is an increasing shift towards artificial intelligence and radiogenomics to identify subtle pre-cancerous changes before clinical symptoms emerge [9]. Investing in early detection is therefore not only clinically advantageous but also economically and socially sustainable for global health systems globally.

**1.3 Limitations of Conventional Diagnostic Methods**

Despite significant advancements, traditional cancer diagnostic methods face numerous limitations that contribute to delayed detection and inaccurate characterization of tumors. Conventional imaging techniques such as CT, MRI, ultrasound, and X-ray rely heavily on radiologist interpretation, making them susceptible to human error, observer variability, and fatigue-induced inconsistencies [9]. Subtle lesions or early-stage malignancies, especially in dense breast tissue or complex anatomical areas, may remain undetected until progression occurs [10]. Biopsy remains the gold standard for definitive diagnosis; however, it is invasive, expensive, and associated with risks such as bleeding, infection, and sampling errors [2]. In addition, tissue biopsies capture only a small tumor region, which may not reflect intratumoral heterogeneity. Histopathological evaluation is time-intensive and depends on pathologist expertise, while inter-observer disagreement is common in borderline cases [6]. Furthermore, conventional diagnostics often fail to predict tumor genetics, treatment response, or metastatic potential accurately. In many low-resource settings, limited access to imaging equipment, laboratory facilities, and oncology specialists leads to long delays in diagnosis and treatment initiation [1]. Radiological images alone cannot reveal molecular abnormalities or gene expression patterns, restricting personalized therapy planning. As a result, patients frequently receive generalized treatment rather than precision medicine tailored to their tumor biology [9]. Moreover, manual feature extraction from images lacks standardization and reproducibility. These constraints highlight the urgent need for automated, data-driven diagnostic frameworks capable of integrating imaging, genomic, and clinical data to provide earlier, more accurate, and personalized cancer detection globally.

**1.4 Rise of AI, Computer Vision, and Radiogenomics in Oncology**

Artificial intelligence (AI), particularly deep learning-based computer vision, is revolutionizing cancer diagnostics by overcoming limitations of conventional radiology and pathology [7]. Convolutional neural networks (CNNs), transformers, and hybrid attention models can automatically learn complex spatial features from CT, MRI, PET, mammography, and histopathology images without manual intervention [3]. These algorithms detect subtle abnormalities such as microcalcifications, early lesions, and irregular cellular patterns invisible to human observers [10]. AI enhances diagnostic accuracy, reduces inter-observer variability, and accelerates decision-making in clinical workflows [4]. Beyond imaging, radiogenomics has emerged as a transformative field linking quantitative imaging biomarkers with tumor genomics, transcriptomics, and epigenetics [1]. Radiogenomic frameworks enable non-invasive prediction of molecular subtypes, gene mutations such as EGFR or BRCA1, treatment resistance, and patient survival outcomes [9]. Integrating imaging and genomic data enables precision oncology by providing deeper understanding of tumor biology without repeated biopsies. AI-driven systems can also perform automated tumor segmentation, disease staging, recurrence prediction, and therapy response monitoring [2]. Moreover, cloud computing and federated learning have facilitated multi-institutional collaboration while preserving patient privacy [8]. As hospitals transition to digital pathology and PACS archives, large datasets are becoming available for training robust AI models. However, challenges remain, including data heterogeneity, algorithm bias, lack of interpretability, and regulatory constraints [6]. Despite these barriers, AI and radiogenomics are increasingly viewed as essential tools for early cancer detection and personalized care, marking a shift from subjective interpretation to data-centric decision-making globally.

**1.5 Aim and Scope of the Article**

This article aims to provide a comprehensive review of how deep learning-based computer vision models and radiogenomic integration frameworks are transforming early cancer detection across multimodal medical imaging platforms [5]. It first examines global cancer challenges, emphasizing the need for early diagnosis to improve survival and reduce socioeconomic burdens [1]. The article then explores traditional imaging and diagnostic limitations, setting the foundation for understanding why artificial intelligence-driven solutions are necessary [3]. It evaluates key deep learning architectures, including convolutional neural networks, transformer-based models, and graph neural networks, used in analyzing MRI, CT, PET, ultrasound, and digital pathology images [6]. Special focus is placed on multimodal imaging fusion strategies, such as early, late, and hybrid fusion, that combine spatial, morphological, and functional biomarkers [2]. The integration of imaging and genomic data through radiogenomics is discussed as a mechanism to predict tumor genotypes, treatment response, and molecular subtypes non-invasively [9]. Furthermore, the article addresses challenges such as data heterogeneity, privacy, ethical concerns, interpretability, and clinical translation of AI systems [4]. Finally, future directions including federated learning, digital twins, and regulatory approval pathways for AI-assisted oncology are presented [10]. The overall aim is to provide clinicians, researchers, and biomedical engineers with a structured understanding of current innovations, existing challenges, and future potential in AI-powered cancer diagnostics. This section establishes the roadmap for subsequent discussions, ensuring a seamless transition from theoretical foundations to practical implementations in medical imaging and radiogenomic integration frameworks globally.

## 2. FOUNDATIONS OF CANCER IMAGING AND DIAGNOSTICS

### 2.1 Conventional Imaging Modalities

Conventional imaging modalities form the backbone of cancer diagnosis by providing structural and functional insights into tumor biology [9]. Magnetic resonance imaging (MRI) uses strong magnetic fields and radiofrequency pulses to generate high-resolution soft tissue contrast, making it particularly useful for brain, spinal, breast, and pelvic imaging [14]. Computed tomography (CT) employs rotating X-rays and computerized reconstruction to visualize cross-sectional body images, offering excellent spatial resolution for lung, liver, and bone lesions [11]. Positron emission tomography (PET) combines radioactive tracers such as fluorodeoxyglucose with CT or MRI to assess metabolic activity, hypoxia, and cellular proliferation within tumors [16]. Ultrasound imaging relies on high-frequency sound waves to create real-time images of soft tissues and blood flow, frequently used in breast, liver, and gynecological cancer screening [8]. Each modality provides unique advantages in terms of contrast, radiation exposure, and accessibility, yet none alone delivers fully comprehensive tumor characterization.

Histopathology remains the gold standard for definitive cancer diagnosis through microscopic examination of stained tissue sections [12]. Following biopsy, tissue samples are fixed, sectioned, and analyzed to determine malignancy, tumor grade, and invasion depth [15]. Digital pathology transforms glass slides into high-resolution virtual images that can be archived, shared, and analyzed using computer-based tools [13]. It enables telepathology, automated cell segmentation, and machine learning-assisted classification. Immunohistochemistry enhances this by detecting protein expression patterns to classify molecular tumor subtypes [10]. This method assists in determining hormone receptor status in breast cancers and proliferation markers like Ki-67. Frozen section analysis also provides rapid intraoperative diagnosis but may compromise structural detail. It is still widely used in surgical oncology for quick decisions [14].

### 2.2 Limitations of Traditional Radiology and Pathology

Traditional radiology and pathology face significant challenges that impact timely and accurate cancer diagnosis [15]. Radiologists manually interpret imaging scans, making decisions vulnerable to fatigue, cognitive bias, and varying experience levels [9]. Two specialists may offer different assessments of the same lesion, particularly in mammography and lung nodule detection, resulting in inter-observer variability [13]. Subtle abnormalities may be overlooked in early-stage cancers with low contrast. As illustrated in Figure 1, each imaging modality has inherent trade-offs in resolution, sensitivity, radiation risk, and anatomical detail, making individual techniques insufficient for comprehensive assessment [17].

Pathology experiences similar constraints. Tissue processing errors, staining variability, and subjective tumor grading can lead to diagnostic disagreement [11]. Manual microscopy limits throughput and creates delays in treatment planning within high-volume oncology centers [16]. Additionally, feature extraction from histological slides is performed visually, depending on a pathologist's ability to identify nuclear pleomorphism, mitotic figures, and stromal invasion [10]. This approach lacks quantitative precision and reproducibility. Resource-limited regions experience further delays due to insufficient pathologists and laboratory infrastructure [12]. Without innovation, diagnostic delays will continue to worsen cancer outcomes nationwide [8] each year.

### 2.3 Introduction to Radiomics and Quantitative Imaging

Radiomics emerged to overcome subjective image interpretation by converting medical images into quantitative data through high-dimensional feature extraction [13]. These features capture tumor phenotype by analyzing pixel intensity, spatial variation, and geometric properties beyond human visual capacity [9]. Texture features evaluate heterogeneity using gray-level co-occurrence matrices, entropy, and run-length statistics, while intensity features assess signal variation within lesions [14]. Shape descriptors quantify margin irregularity, sphericity, and surface texture, often correlated with malignant behavior [16]. Together, these handcrafted biomarkers provide non-invasive insight into tumor aggression, prognosis, and therapy response.

Radiomics enabled early associations between imaging biomarkers and survival outcomes; however, manual feature engineering introduces challenges including segmentation variability, poor reproducibility, and dependence on imaging protocols [12]. Changes in scanner type, voxel size, and noise significantly alter radiomic values [15]. Furthermore, handcrafted features fail to capture complex hierarchical patterns in cancer imaging data, limiting predictive accuracy for highly heterogeneous tumors [17]. These limitations encouraged adoption of machine learning classifiers and subsequently deep learning methods. Deep neural networks automatically learn abstract features directly from raw MRI,

CT, or histopathology images without manual design [8]. This transition from handcrafted radiomics to deep learning represents a major advancement toward scalable, objective, and precise cancer detection across clinical practice.

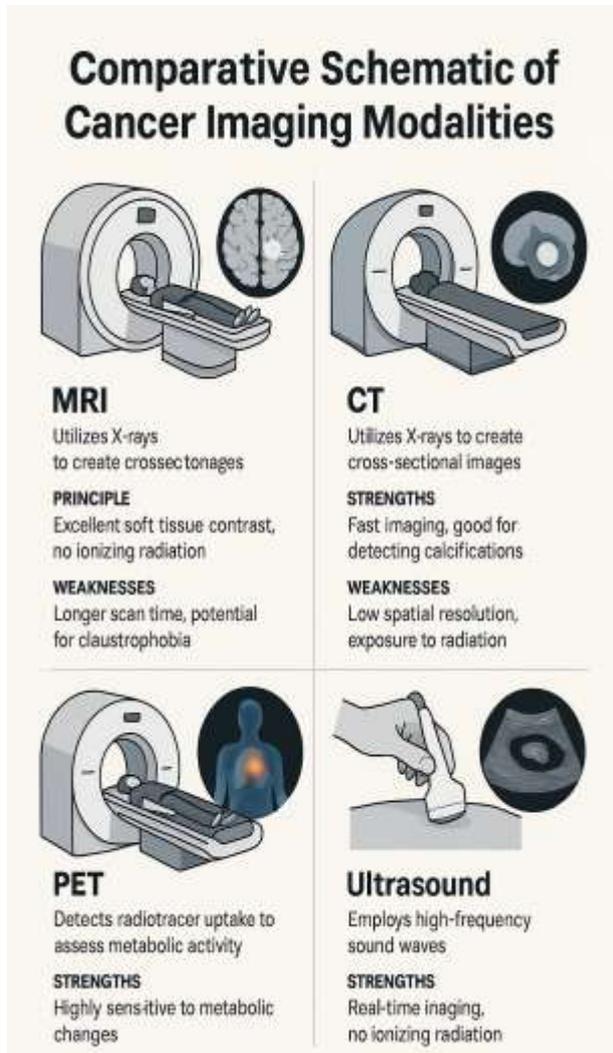

Figure 1: Comparative schematic of cancer imaging modalities [6]

# 3. DEEP LEARNING AND COMPUTER VISION IN CANCER DETECTION
## 3.1 Evolution from Machine Learning to Deep Learning

Machine learning has long been used in cancer diagnosis to classify tumors using handcrafted features extracted from imaging data [18]. However, traditional algorithms such as support vector machines and random forests rely heavily on domain expertise for feature engineering and cannot autonomously learn hierarchical representations from raw data [15]. Deep learning emerged as a transformative solution, allowing artificial neural networks to automatically learn complex spatial, textural, and morphological features from medical images [19]. Convolutional neural networks (CNNs) revolutionized computer vision by enabling automated detection of tumors, microcalcifications, and metastases from MRI, CT, PET, and histopathological images [22]. Autoencoders contributed through unsupervised learning, allowing dimensionality reduction and anomaly detection in unlabeled medical datasets [17]. More recently, transformers, originally developed for natural language processing, introduced self-attention mechanisms that capture long-range dependencies in imaging and genomic data [23]. These models outperform conventional CNNs in tumor segmentation, radiogenomic prediction, and multi-modal fusion tasks. Deep learning has shifted cancer imaging from handcrafted to data-driven feature extraction, improving sensitivity, reproducibility, and robustness across clinical workflows [25]. Its evolution marks the beginning of scalable precision oncology by integrating imaging, genomics, and clinical metadata for early cancer detection [21].

## 3.2 Core Architectures

Deep learning architectures used in cancer imaging vary in complexity, accuracy, and interpretability [20]. Convolutional Neural Networks (CNNs) form the foundation of most radiological and histopathological models. They consist of convolutional layers that learn spatial hierarchies of features such as edges, masses, and cellular atypia from imaging data [16]. CNNs are widely applied in tumor detection, breast lesion classification, and lung nodule screening [24]. However, standard CNNs struggle with gradient vanishing in deeper networks, leading to the development of advanced architectures.

Residual Networks (ResNets) introduced skip connections that allow gradients to bypass certain layers, enabling deeper networks exceeding 100 layers without performance degradation [18]. DenseNets further improved efficiency by connecting each layer to every other layer through feature concatenation, enhancing feature reuse and reducing parameter redundancy [22]. These networks have demonstrated superior accuracy in classifying malignancies in mammography and segmenting brain tumors.

Vision Transformers (ViT) represent a paradigm shift by dividing images into patches and processing them using self-attention mechanisms rather than convolutions [19]. This enables the model to capture long-range dependencies and global context, beneficial for heterogeneous tumors. Swin Transformers enhance ViT by incorporating hierarchical windows and shifted attention, improving computational efficiency for high-resolution medical images [25].

These models serve as the backbone for radiomics and radiogenomics analysis. Table 1 summarizes commonly used metrics for evaluating model performance, including accuracy, sensitivity, and Dice similarity [17]. Their successful deployment supports scalable clinical decision-making across oncology.

## 3.3 Training Data, Labeling, and Preprocessing

Deep learning models require large volumes of annotated imaging data to achieve clinical-grade accuracy [15].

However, medical datasets are often limited due to privacy restrictions, scarcity of labeled images, and variability in imaging protocols across institutions [23]. Expert radiologists and pathologists provide manual annotations for tumor boundaries, tissue regions, and malignancy grades, but this process is time-consuming and prone to observer bias [19]. To improve model generalizability, preprocessing steps are essential.

Data augmentation techniques such as rotation, flipping, zooming, and elastic deformation increase dataset variability by synthetically generating new training samples [21]. These ensure the model is robust to anatomical and positional variations across patients [24]. Normalization adjusts pixel intensity distributions to reduce scanner-related discrepancies between MRI or CT images [20]. Standardization of voxel spacing and resolution improves feature consistency.

Class imbalance poses a major challenge, especially in datasets where malignant cases are significantly fewer than normal samples [18]. Oversampling, focal loss functions, and synthetic data generation techniques such as SMOTE and GANs are used to prevent models from biasing toward majority classes [25]. These strategies collectively ensure accurate and reproducible training of AI systems for cancer detection.

### 3.4 Evaluation Metrics

Evaluating deep learning-based cancer detection models requires quantitative metrics that assess diagnostic accuracy, segmentation performance, and clinical relevance [23]. Commonly reported parameters include sensitivity, specificity, precision, accuracy, and F1-score. Sensitivity reflects a model's ability to correctly identify cancer-positive cases, while specificity measures accurate detection of non-cancer cases [16]. High sensitivity is critical to avoid missed diagnoses, especially in early-stage cancer screening [19].

The Receiver Operating Characteristic Area Under the Curve (ROC-AUC) provides a comprehensive measure of classification performance across various threshold settings [21]. A higher AUC indicates better discriminatory capability between malignant and benign lesions [25]. For segmentation-based tasks such as tumor boundary detection in MRI or histopathology, overlap-based metrics are essential.

The Dice Similarity Coefficient (DSC) measures spatial agreement between predicted and ground-truth tumor masks, with values closer to 1 indicating superior segmentation accuracy [17]. Intersection over Union (IoU) is also widely used for lesion localization [22]. In radiogenomic frameworks, metrics extend to concordance indices for survival prediction and correlation scores for gene–image associations [20].

Table 1 provides a structured summary of metrics commonly applied in imaging-based cancer models. Accurate reporting of these indicators ensures standardization and enables meaningful comparison between AI systems across institutions [18]. Furthermore, evaluation should incorporate clinician-in-the-loop validation and external multi-center datasets to reduce bias [24]. This holistic approach ensures AI models support safe, reliable, and adoptable clinical decision-making across oncology.

**Table 1: Common Evaluation Metrics Used in Imaging-Based Cancer Detection Models**

| Metric Name | Formula / Description | Purpose in Cancer Imaging | Clinical Relevance |
|---|---|---|---|
| Accuracy | (TP + TN) / (TP + TN + FP + FN) | Measures overall correct predictions made by the model. | Useful for balanced datasets but misleading when cancer cases are rare. |
| Sensitivity (Recall / True Positive Rate) | TP / (TP + FN) | Ability to correctly detect cancer-positive cases. | Critical to avoid missed diagnoses in early cancer detection. |
| Specificity (True Negative Rate) | TN / (TN + FP) | Ability to correctly identify cancer-free individuals. | Reduces false alarms and unnecessary biopsies or imaging. |
| Precision (Positive Predictive Value) | TP / (TP + FP) | Percentage of predicted cancer cases that are truly cancer. | Important for reducing false positives and patient anxiety. |
| F1-Score | 2 × (Precision × Recall) / (Precision + Recall) | Harmonic mean of precision and sensitivity. | Balanced indicator in imbalanced cancer datasets. |
| ROC-AUC (Receiver Operating Characteristic – Area Under Curve) | Probability that a model ranks a random positive higher than a random negative. | Evaluates diagnostic discrimination capability across thresholds. | Widely used for model comparison in oncology. |
| Dice Similarity Coefficient (DSC) | 2 × | X ∩ Y | / ( |  |
| IoU (Intersection | | X ∩ Y | / |

| Metric Name | Formula / Description | Purpose in Cancer Imaging | Clinical Relevance |
|---|---|---|---|
| over Union) | | | |
| Mean Absolute Error (MAE) | Mean of | Predicted − True | |
| Concordance Index (C-Index) | Probability model predicts survival ranking correctly | Evaluates survival prediction and risk stratification accuracy. | Important in radiogenomic prognostic modeling. |

## 4. MULTIMODAL MEDICAL IMAGING INTEGRATION

### 4.1 Rationale for Multimodal Fusion

Multimodal fusion in cancer imaging integrates complementary information from multiple diagnostic techniques to enhance early detection, staging, and treatment planning [26]. Single imaging modalities offer valuable but incomplete insights into tumor biology. MRI provides excellent soft-tissue contrast and detailed anatomical structures, while CT delivers precise information on bone involvement and tumor morphology [24]. PET contributes metabolic and functional data, highlighting glucose uptake patterns associated with tumor aggressiveness [27]. Histopathology and digital pathology enable cellular-level visualization and assessment of tissue architecture, serving as the gold standard for diagnosis [23]. However, each modality alone is limited by resolution, noise, or lack of molecular insight.

By combining modalities, multimodal fusion improves diagnostic sensitivity and specificity, reduces false negatives, and enables non-invasive prediction of tumor progression [29]. Integration allows radiologists and oncologists to correlate spatial, metabolic, and microscopic information from the same patient, enabling more precise tumor margin identification [25]. This holistic approach supports precision oncology by aligning imaging features with underlying tumor biology and treatment response patterns [30]. Consequently, multimodal imaging is increasingly used in radiogenomics, radiation therapy planning, and surgical navigation. It forms the foundation for AI-driven diagnostic pipelines that analyze cross-domain data for more accurate cancer detection and prognosis globally.

### 4.2 Fusion Techniques

Multimodal fusion techniques are essential for integrating information from MRI, CT, PET, histopathology, and genomic data to enhance tumor characterization [28]. These strategies can be classified into early, late, and hybrid fusion approaches based on the stage at which data is combined.

Early fusion integrates raw imaging data or features extracted from different modalities at the input stage before being processed by a model [23]. This approach preserves spatial and textural relationships but is sensitive to misalignment and varying resolution across modalities [26]. In contrast, late fusion combines independently processed features or classification outputs from separate models, merging them at the decision-making stage [31]. Although more flexible, late fusion may ignore deep interdependencies between modalities.

Hybrid fusion combines the strengths of both early and late fusion by integrating features at multiple network layers, enabling both low-level and high-level feature correlation [27]. This method achieves superior accuracy in tasks such as tumor segmentation and metastasis prediction.

More advanced architectures include multi-stream CNNs, where each modality is processed through separate convolutional streams before feature concatenation [25]. These models capture unique modality-specific representations. Attention-based models, such as cross-modal attention networks, dynamically weigh the contribution of each modality depending on relevance to the tumor region [32]. Transformers have also been adapted to multimodal imaging due to their ability to model long-range dependencies across heterogeneous inputs [30].

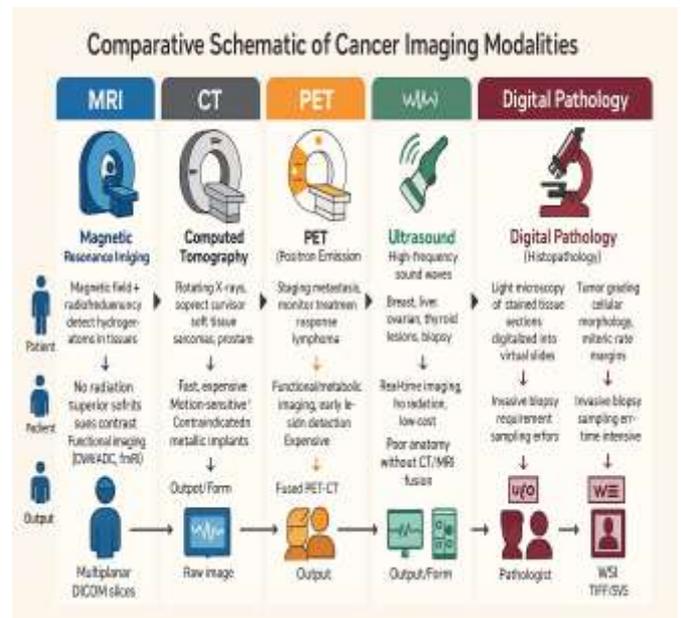

Figure 2 Multimodal image fusion pipeline

Figure 2 illustrates a typical multimodal image fusion pipeline involving image preprocessing, modality registration, feature extraction, fusion layers, and classification or segmentation outputs [29]. These frameworks underpin radiogenomic models integrating imaging with genetic expression profiles.

**4.3 Applications to Organ-Specific Cancers**

Multimodal fusion has demonstrated significant improvements in diagnosis and prognosis across multiple cancer types [33]. In breast cancer, MRI and mammography fusion provides enhanced sensitivity for detecting early lesions in dense breast tissue [28]. PET-CT is widely used for identifying axillary lymph node metastasis and assessing treatment response in neoadjuvant therapy [24]. MRI combined with digital pathology further enables prediction of molecular subtypes such as HER2-positive or triple-negative breast cancer [27].

In lung cancer, integrating PET and CT is the clinical standard for staging and detecting distant metastases [30]. Multistream CNNs trained on fused PET-CT data outperform single-modality models in nodule classification and malignancy prediction [25]. MRI fusion supports evaluation of mediastinal invasion in advanced cases [26].

Brain glioma diagnosis benefits from combining MRI sequences (T1, T2, FLAIR) with PET to assess tumor infiltration, edema, and metabolic activity [29]. Fusion models can distinguish between low-grade and high-grade gliomas more accurately than MRI alone [31]. Radiogenomic studies have linked MRI-PET fusion features with IDH mutation status and MGMT promoter methylation [23].

Colorectal cancer applications involve integrating CT colonography and PET to detect primary tumors and liver metastasis [32]. Histopathology fusion aids in tumor budding and lymphovascular invasion assessment [28]. These organ-specific examples highlight how multimodal imaging improves sensitivity, reduces diagnostic delays, and supports precision therapy pathways across oncology [27].

**4.4 Case Study Transition**

To illustrate the real-world application of multimodal fusion, the next section presents a case study focusing on glioblastoma, an aggressive brain cancer characterized by high heterogeneity and poor survival outcomes [25]. In this case study, MRI, PET, and genomic data integration are examined to demonstrate how deep learning models detect early microstructural changes and predict genetic mutations non-invasively [31]. The workflow mirrors the architecture shown in Figure 2, including preprocessing, modality registration, feature extraction, and hybrid fusion strategies [29]. This transition sets the stage for practical insights into clinical translation of AI-based multimodal cancer detection systems.

## 5. RADIOGENOMICS AND AI-DRIVEN GENOMIC INTEGRATION

**5.1 Introduction to Radiogenomics**

Radiogenomics is an emerging discipline that connects medical imaging features with genomic, transcriptomic, or epigenetic alterations in tumors, enabling non-invasive prediction of molecular profiles [33]. The core hypothesis is that tumor genetics influence morphology, metabolism, texture, and microenvironment characteristics that can be captured through imaging phenotypes [36]. For example, tumor heterogeneity observed on MRI or CT may reflect variations in angiogenesis, hypoxia, or cellular proliferation driven by specific gene mutations [31]. Radiogenomics offers a superior alternative to traditional biopsies, which are invasive, limited to sampled regions, and unable to capture intratumoral heterogeneity [37]. By learning imaging biomarkers associated with genetic changes, radiogenomic models can predict gene expression patterns, mutation status, and treatment sensitivity without requiring tissue extraction [30]. This approach is particularly valuable for brain, breast, lung, and liver cancers where biopsies are risky or insufficient. Moreover, radiogenomics supports personalized oncology by integrating imaging, pathology, and genomic data into predictive algorithms that guide therapy decisions [35]. The integration of artificial intelligence and deep learning further enhances radiogenomic workflows by enabling automated segmentation, feature extraction, and high-dimensional data fusion [38]. Consequently, radiogenomics represents a transformative step toward precision medicine in cancer detection and treatment.

**5.2 Gene–Imaging Correlation Models**

Radiogenomic correlation models aim to uncover relationships between imaging phenotypes and gene expression patterns using population datasets and machine learning algorithms [34]. The Cancer Genome Atlas (TCGA) and The Cancer Imaging Archive (TCIA) are primary resources containing paired genomic and imaging data for multiple cancer types, including glioblastoma, breast, lung, and renal cancers [32]. These datasets enable training of prediction models to link tumor texture, shape, and perfusion parameters with gene mutations and transcriptomic signatures [36].

In breast cancer, radiogenomic studies predict BRCA1/BRCA2 mutation status using dynamic contrast-enhanced MRI features, including lesion enhancement kinetics and morphological irregularity [39]. Similarly, lung cancer studies correlate CT-derived radiomic signatures with EGFR and KRAS mutations to determine targeted therapy eligibility [33]. These models can predict EGFR mutations with high accuracy by analyzing spiculated margins, ground-glass opacities, and metabolic PET uptake [37]. In glioblastoma, MRI features such as necrotic core volume, peritumoral edema, and contrast-enhancing ring patterns have been linked to IDH mutation and MGMT promoter methylation [35]. These findings enable non-invasive molecular subtyping.

Radiogenomic pipelines typically involve preprocessing, segmentation, feature extraction, and predictive modeling. Figure 3 illustrates a typical radiogenomic integration architecture, combining MRI, CT, or PET features with multi-omics data using deep learning fusion networks [31]. Despite promising results, variability in imaging protocols and sample sizes across institutions can affect reproducibility [38]. Therefore, robust validation and standardized imaging acquisition are essential for reliable gene–imaging correlations in clinical practice [30].

**5.3 Deep Learning Architectures in Radiogenomics**

Deep learning has advanced radiogenomics by enabling automatic feature discovery from imaging and genomic data without manual engineering [32]. Among the most impactful architectures are Graph Neural Networks (GNNs), which model relationships between interconnected biological and imaging features. GNNs represent genes, radiomic descriptors, and clinical attributes as nodes, while edges encode functional interactions or spatial relationships within tumors [36]. In glioblastoma, GNNs have been used to predict IDH mutation by combining MRI-derived spatial voxels with gene co-expression networks [34]. This architecture captures complex gene–imaging interactions and tumor microenvironment patterns often missed by traditional models [35].

Transformer-based fusion encoders have also revolutionized radiogenomics through self-attention mechanisms that weigh the importance of each imaging and genomic feature relative to others [39]. These models divide imaging data into patches and genomic data into token embeddings, which are processed through multi-head attention layers to capture long-range dependencies [31]. Multimodal transformers have successfully predicted BRCA mutation status in breast cancer and EGFR mutations in lung cancer using MRI and CT radiomic signatures combined with gene expression profiles [33].

Hybrid models integrating CNNs, autoencoders, and transformers have improved lesion-level mutation prediction in hepatocellular carcinoma and colorectal cancer [38]. These models support early detection and therapy selection by producing probability maps linking tumor regions to genomic alterations. Table 2 summarizes key imaging biomarkers and their associated genetic mutations across common cancers, demonstrating their diagnostic importance [30]. However, challenges such as limited annotated data, interpretability, and computational complexity must be addressed before full clinical adoption [40]. Federated learning and transfer learning are emerging to overcome data privacy and scalability obstacles [37].

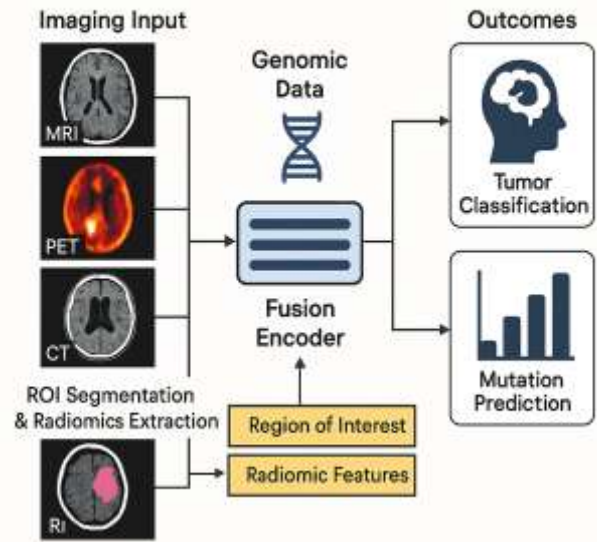

**Figure 3: Radiogenomic integration model architecture [22]**

**5.4 Radiogenomic Biomarkers and Clinical Relevance**

Radiogenomic biomarkers play a critical role in precision oncology by connecting imaging characteristics with underlying genomics to inform diagnosis, prognosis, and treatment guidance [32]. Imaging biomarkers such as peritumoral edema, tumor heterogeneity, irregular margins, and enhancement intensity have been associated with specific gene mutations including EGFR, BRCA, KRAS, IDH, and TP53 [38]. These biomarkers allow clinicians to non-invasively infer molecular subtypes, reducing the need for repeated biopsies in inaccessible tumors like gliomas [36]. Additionally, radiogenomic models predict therapy response to targeted treatments, immunotherapy, and chemoradiation by analyzing metabolic PET activity, perfusion MRI, and gene expression patterns [33].

Clinical relevance extends to survival prediction, where integration of imaging phenotypes with genomic markers produces more accurate prognostic scores than conventional TNM staging systems [39]. In radiation oncology, radiogenomics assists in adaptive radiotherapy planning by monitoring spatial gene expression across tumor regions [37]. Table 2 provides a structured overview of imaging biomarkers and their corresponding genetic mutations in breast, lung, brain, and colorectal cancers [34]. Despite progress, clinical translation requires multicenter validation, ethical data governance, and integration with electronic health records [35]. Radiogenomics is evolving into a cornerstone of personalized medicine in oncology.

**Table 2: Imaging Biomarkers vs Genetic Mutations in Common Cancers**

| Cancer Type | Imaging Biomarker | Modality | Associated Genetic Mutation / Molecular Marker | Clinical Relevance |
|---|---|---|---|---|
| Breast Cancer | Irregular spiculated margins, heterogeneous enhancement | Mammography / MRI | **BRCA1/BRCA2**, **HER2**, **PIK3CA** | BRCA mutations linked to triple-negative tumors and aggressive phenotypes; HER2 amplification predicts trastuzumab response. |
| | Rim enhancement and rapid wash-in/wash-out kinetics | Dynamic Contrast-Enhanced MRI (DCE-MRI) | **TP53**, **BRCA1** | Indicates high angiogenesis and poor prognosis in basal-like subtypes. |
| Lung Cancer | Ground-glass opacities, spiculated nodules | CT | **EGFR**, **KRAS**, **ALK** | EGFR mutations common in non-smokers; associated with targeted therapy eligibility (gefitinib, erlotinib). |
| | High SUV uptake in lesions | PET-CT | **KRAS**, **TP53** | Metabolic activity correlates with tumor aggressiveness and treatment resistance. |
| Brain (Glioblastoma) | Necrotic core with ring enhancement, peritumoral edema | MRI (T1/T2/FLAIR) | **IDH1/2**, **MGMT promoter methylation**, **TERT** | IDH mutation indicates better prognosis; MGMT methylation predicts temozolomide response. |
| | Reduced ADC and increased rCBV (relative cerebral blood volume) | Diffusion / Perfusion MRI | **EGFRvIII**, **PTEN** | Indicates proliferation index and invasion potential. |
| Colorectal Cancer | Irregular bowel wall thickening, mucosal disruption | CT colonography | **KRAS**, **NRAS**, **BRAF** | Mutations influence EGFR inhibitor response (cetuximab, panitumumab). |
| | Liver hypovascular metastasis | MRI / PET | **APC**, **TP53** | Suggests metastatic progression linked to WNT pathway dysregulation. |

## 6. CHALLENGES, ETHICS, AND IMPLEMENTATION BARRIERS
### 6.1 Technical Challenges

Despite significant advances, AI-driven cancer detection systems face critical technical barriers, particularly in data scarcity, image noise, and heterogeneity across imaging modalities and institutions [41]. Deep learning models require large, annotated datasets to achieve generalizable performance, yet medical data is limited due to privacy laws, proprietary restrictions, and costly labeling processes requiring expert oncologists and radiologists [38]. Additionally, heterogeneous imaging protocols, scanner types, and reconstruction settings across hospitals introduce variability that affects model reproducibility [44]. Differences in pixel resolution, contrast injection timing, and slice thickness can degrade model accuracy when applied to external datasets [39].

Noise induced by motion artifacts, low-dose CT protocols, or poor histopathological staining also contributes to misclassification errors [42]. In pathology, variations in slide preparation, staining intensity, and tissue folding create inconsistencies in digital image analysis [40]. Small sample sizes for rare cancers further complicate model training, increasing the risk of overfitting and bias [45]. While data augmentation and domain adaptation techniques mitigate some limitations, they cannot fully compensate for real-world diversity in patient populations [43]. Multi-center collaboration, federated learning, and standardized imaging protocols are therefore essential to develop clinically robust AI models capable of reliable deployment across diverse healthcare ecosystems [46].

### 6.2 Interpretability and Explainable AI

AI models used in oncology are often considered "black boxes," providing accurate predictions without explaining the underlying decision-making process [47]. This lack of transparency limits trust among clinicians who must justify medical decisions to patients and regulatory bodies [40]. Explainable AI (XAI) techniques such as Gradient-Weighted Class Activation Mapping (Grad-CAM) and saliency maps help visualize which regions of an image contribute most to the model's predictions [39]. Grad-CAM overlays heatmaps on MRI, CT, or histopathology images to highlight tumor regions responsible for classification outputs [44]. Saliency maps compute gradients with respect to input pixels, offering insight into important spatial features learned by neural networks [42].

These techniques improve accountability and help detect erroneous predictions caused by artifacts or noise. Figure 4 illustrates key interpretability challenges in AI-based cancer imaging, including mislocalized attention and model bias across demographic groups [46]. For radiogenomics, interpretability is even more critical due to combined imaging and genomic inputs [41]. However, current XAI methods are limited by inconsistency and lack of clinical validation. Future systems must incorporate interpretable architectures and uncertainty quantification to gain regulatory approval and clinician confidence in real-world environments [45].

### 6.3 Ethical and Legal Concerns

The integration of AI and radiogenomics in cancer care introduces ethical and legal concerns related to patient privacy, informed consent, and data governance [38]. Genomic data is uniquely sensitive, as it carries hereditary information that can impact family members and increase risks of discrimination by insurers or employers [41]. Inadequate anonymization can allow patient re-identification from imaging-genomic datasets [43]. Ethical AI deployment requires transparent data usage policies, encryption, and compliance with regulations such as GDPR and HIPAA [46]. Patients must be informed about how their imaging, pathology, and genomic data will be used in training AI systems, enabling meaningful consent rather than passive agreement [40]. Algorithmic bias is another concern, as models trained predominantly on data from Western populations may underperform in African or Asian groups [44]. Addressing these risks is essential for ethically sustainable AI-enabled cancer diagnostics [47].

### 6.4 Deployment Barriers in Low-Resource Settings

Deploying AI-based cancer imaging tools in low-resource settings remains challenging due to inadequate infrastructure, limited funding, and shortage of trained specialists [42]. Many hospitals lack high-speed internet, GPU computing devices, and digital pathology scanners required for AI workflows [39]. Power instability, outdated imaging equipment, and absence of picture archiving and communication systems (PACS) hinder seamless implementation [43]. Licenses for cloud platforms and proprietary AI tools are often unaffordable [45]. Workforce limitations include insufficient data scientists, radiologists, and biomedical engineers [46]. Sustainable deployment requires low-cost AI models, offline-compatible systems, and local capacity building to bridge global healthcare disparities [44].

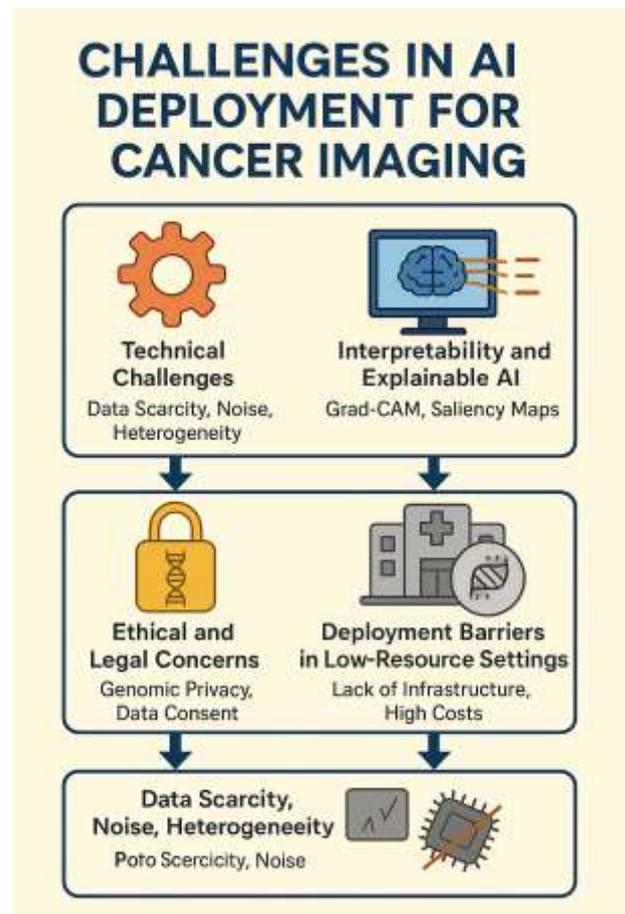

Figure 4: Challenges in AI deployment for cancer imaging

# 7. FUTURE DIRECTIONS AND CLINICAL TRANSLATION
## 7.1 Federated Learning and Collaborative AI

Federated learning has emerged as a transformative approach for training artificial intelligence models across multiple hospitals without requiring data centralization [46]. Instead of sending sensitive imaging or genomic data to a shared server, federated learning enables hospitals to train models locally and only share updated parameters with a central aggregator [42]. This preserves patient privacy, complies with data protection regulations, and enhances collaboration among institutions with diverse imaging protocols and patient demographics [49]. Such decentralized learning significantly improves model generalizability by incorporating population-specific variations across regions and healthcare systems [44].

For oncology applications, federated learning has been deployed for lung nodule detection, breast cancer mammogram analysis, and brain tumor segmentation using MRI datasets from international institutions [50]. Blockchain technology and differential privacy further secure parameter transmission and prevent data leakage during communication [47]. Collaborative AI frameworks also allow inclusion of genomic and pathology data while adhering to informed consent protocols and ethical standards [45]. Despite its benefits, federated learning faces challenges such as data heterogeneity, non-uniform imaging quality, and bandwidth limitations in low-resource settings [48]. Nevertheless, it provides a feasible pathway toward scalable and ethically compliant AI deployment in cancer diagnostics globally.

## 7.2 Digital Twins and Precision Oncology

Digital twins represent virtual replicas of individual patients, integrating medical imaging, genomics, clinical records, and treatment responses to simulate cancer progression in real time [43]. These computational models allow oncologists to test multiple therapy strategies in a digital environment before applying them to the patient, minimizing trial-and-error in treatment planning [49]. In oncology, digital twins incorporate tumor growth models, radiomic signatures, and molecular alterations to predict therapeutic outcomes for surgery, chemotherapy, or immunotherapy [46]. They enable precision oncology by tailoring interventions based on patient-specific biological and anatomical characteristics rather than generalized treatment guidelines [45].

Using MRI, CT, or PET data, digital twins can map tumor microenvironment changes and vascular alterations at each treatment stage [42]. Integration with radiogenomic frameworks allows simulation of genetic mutation effects such as EGFR or BRCA1 alterations on tumor metabolism and drug sensitivity [50]. These simulations can also assess radiation dose distribution for adaptive radiotherapy. However, implementation demands high computational power, standardized imaging protocols, and longitudinal data collection [44]. Ethical challenges include informed consent and potential misuse of predictive models without clinical supervision [47]. Despite these challenges, digital twins are poised to redefine cancer treatment personalization and predictive oncology.

## 7.3 Regulatory Approval and Clinical Trials

For AI-based cancer detection tools to transition from research to clinical practice, they must undergo rigorous validation, regulatory approval, and multi-center clinical trials [48]. The United States Food and Drug Administration (FDA) categorizes AI systems as Software as a Medical Device (SaMD) and requires demonstration of analytical validity, clinical performance, and patient safety [42]. The FDA's 510(k) and De Novo pathways have approved AI tools for mammography, lung nodule detection, and digital pathology analysis [50]. In Europe, the European Medicines Agency (EMA) enforces similar standards under the Medical Device Regulation (MDR), emphasizing algorithm transparency, cybersecurity, and post-market surveillance [46].

Clinical trials for AI in oncology require prospective validation, randomized comparative studies, and external dataset testing across diverse populations [47]. Regulatory frameworks increasingly support adaptive AI systems, provided that model updates remain documented and auditable [49]. Additionally, ethical boards require robust consent mechanisms and data governance in radiogenomic-based clinical trials [45]. Industry collaboration with hospitals and academic institutions ensures compliance with legal, ethical, and technical standards [44]. Successful approval and deployment of AI tools depend on interdisciplinary coordination across software developers, oncologists, radiologists, regulatory bodies, and policymakers [43].

# 8. CONCLUSION
## 8.1 Summary of Contributions

This article explored how deep learning-driven computer vision, multimodal imaging, and radiogenomic integration frameworks are transforming early cancer detection. It outlined the evolution from conventional diagnostics to AI-based systems capable of non-invasively predicting tumor genetics, improving sensitivity, and reducing diagnostic delays. Key contributions included a detailed review of fusion techniques, radiogenomic models, digital pathology, explainable AI, and federated learning. It further highlighted the integration of MRI, CT, PET, and genomic data in precision oncology. Technical challenges, ethical concerns, and infrastructure barriers were critically discussed to underscore the readiness and limitations of AI-driven oncology.

## 8.2 Multidisciplinary Advancements

The advancement of radiogenomic AI is a convergence of radiology, oncology, genomics, biomedical engineering, and data science. Radiomics and digital pathology have enabled the extraction of quantitative biomarkers, while deep learning architectures such as CNNs, transformers, and graph neural networks allow automated interpretation of high-dimensional medical data. Genomic sequencing complements imaging by revealing molecular alterations that influence treatment response. Collaborative efforts between clinicians, AI researchers, bioinformaticians, and regulatory authorities are driving clinically validated diagnostic models. Additionally, cloud computing, edge AI, and federated learning enable secure cross-institutional training, enhancing generalizability and reducing data privacy risks in global cancer research.

### 8.3 Future of Radiogenomic AI in Precision Medicine

The future of radiogenomic AI lies in fully integrated, clinically deployable systems that provide personalized diagnosis, treatment selection, and therapy monitoring. Digital twins will simulate patient-specific tumor evolution, enabling adaptive interventions and predictive modeling. Federated learning will facilitate global AI development without compromising patient data privacy. Explainable AI will build clinician trust through transparent decision-making. Integration with electronic health records and real-time hospital workflows will make AI a routine clinical tool rather than a research prototype. Ultimately, radiogenomic AI will shift cancer care from reactive to predictive, enabling earlier detection, minimally invasive treatment, and improved survival outcomes in precision medicine.

## 9. REFERENCE

1. Gou X, Feng A, Feng C, Cheng J, Hong N. Imaging genomics of cancer: a bibliometric analysis and review. Cancer Imaging. 2025 Mar 4;25(1):24.
2. Oni D. Hospitality industry resilience strengthened through U.S. government partnerships supporting tourism infrastructure, workforce training, and emergency preparedness. *World Journal of Advanced Research and Reviews.* 2025;27(3):1388–1403. doi:https://doi.org/10.30574/wjarr.2025.27.3.3286
3. Hussain S, Lafarga-Osuna Y, Ali M, Naseem U, Ahmed M, Tamez-Peña JG. Deep learning, radiomics and radiogenomics applications in the digital breast tomosynthesis: a systematic review. BMC bioinformatics. 2023 Oct 26;24(1):401.
4. Solarin A, Chukwunweike J. Dynamic reliability-centered maintenance modeling integrating failure mode analysis and Bayesian decision theoretic approaches. *International Journal of Science and Research Archive*. 2023 Mar;8(1):136. doi:10.30574/ijsra.2023.8.1.0136.
5. Sui D, Guo M, Ma X, Baptiste J, Zhang L. Imaging biomarkers and gene expression data correlation framework for lung cancer radiogenomics analysis based on deep learning. IEEE Access. 2021 Apr 6;9:125247-57.
6. Oni Daniel. The U.S. government shapes hospitality standards, tourism safety protocols, and international promotion to enhance competitive global positioning. *Magna Scientia Advanced Research and Reviews.* 2023;9(2):204-221. doi:https://doi.org/10.30574/msarr.2023.9.2.0163
7. Guo Y, Li T, Gong B, Hu Y, Wang S, Yang L, Zheng C. From Images to Genes: Radiogenomics Based on Artificial Intelligence to Achieve Non-Invasive Precision Medicine in Cancer Patients. Advanced Science. 2025 Jan;12(2):2408069.
8. Boehm KM, Khosravi P, Vanguri R, Gao J, Shah SP. Harnessing multimodal data integration to advance precision oncology. Nature Reviews Cancer. 2022 Feb;22(2):114-26.
9. Braman N, Gordon JW, Goossens ET, Willis C, Stumpe MC, Venkataraman J. Deep orthogonal fusion: multimodal prognostic biomarker discovery integrating radiology, pathology, genomic, and clinical data. InInternational conference on medical image computing and computer-assisted intervention 2021 Sep 21 (pp. 667-677). Cham: Springer International Publishing.
10. Temiloluwa Evelyn Olatunbosun, and Cindy Chinonyerem Iheanetu. 2025. "Data-Driven Insights into Maternal and Child Health Inequalities in the U.S". *Current Journal of Applied Science and Technology* 44 (8):98–110. https://doi.org/10.9734/cjast/2025/v44i84593.
11. Steyaert S, Pizurica M, Nagaraj D, Khandelwal P, Hernandez-Boussard T, Gentles AJ, Gevaert O. Multimodal data fusion for cancer biomarker discovery with deep learning. Nature machine intelligence. 2023 Apr;5(4):351-62.
12. Durowoju ES, Olowonigba JK. Machine learning-driven process optimization in semiconductor manufacturing: a new framework for yield enhancement and defect reduction. *[Journal name unavailable]*. 2025;6:1-?. doi:10.55248/gengpi.6.0725.2579.
13. Liu Z, Duan T, Zhang Y, Weng S, Xu H, Ren Y, Zhang Z, Han X. Radiogenomics: a key component of precision cancer medicine. British Journal of Cancer. 2023 Sep 21;129(5):741-53.
14. Amanna A. *Deploying next-generation artificial intelligence ecosystems for real-time biosurveillance, precision health analytics and dynamic intervention planning in life science research.* Magna Scientia Advanced Biology and Pharmacy. 2025;16(1):38-54. doi:10.30574/msabp.2025.16.1.0066
15. Wang X, Li BB. Deep learning in head and neck tumor multiomics diagnosis and analysis: review of the literature. Frontiers in Genetics. 2021 Feb 10;12:624820.
16. Abdikenov B, Zhaksylyk T, Shortanbaiuly O, Orazayev Y, Makhanov N, Karibekov T, Suvorov V, Imasheva A, Zhumagozhayev K, Seitova A. Future of Breast Cancer Diagnosis: A Review of DL and ML Applications and Emerging Trends for Multimodal Data. IEEE Access. 2025 Jul 2.
17. Temiloluwa Evelyn Olatunbosun, and Cindy Chinonyerem Iheanetu. 2025. "Bridging the Gap: Community-Based Strategies for Reducing Maternal and Child Health Disparities in the U.S". *Current Journal of Applied Science and Technology* 44 (8):111–120. https://doi.org/10.9734/cjast/2025/v44i84594.
18. Kumar R, Sporn K, Khanna A, Paladugu P, Gowda C, Ngo A, Jagadeesan R, Zaman N, Tavakkoli A. Integrating Radiogenomics and Machine Learning in Musculoskeletal Oncology Care. Diagnostics. 2025 May 29;15(11):1377.

19. Derera R. Machine learning-driven credit risk models versus traditional ratio analysis in predicting covenant breaches across private loan portfolios. *International Journal of Computer Applications Technology and Research*. 2016;5(12):808-820. doi:10.7753/IJCATR0512.1010.

20. Hou Z, Zhao Z, Zhu D, Pan M, Zhou Y, Ge Q. Advances in Artificial Intelligence-Based Radiomultiomics for Assisting Cancer Diagnosis and Treatment. In2025 International Symposium on Intelligent Robotics and Systems (ISoIRS) 2025 Jun 13 (pp. 01-08). IEEE.

21. Takuro Kehinde Ojadamola. Analyzing Intellectual Property Rights Adaptation to Artificial Intelligence-Created Works and Automated Innovation in the Global Knowledge Economy. *International Journal of Computer Applications Technology and Research*. 2021;10(12):414-424. doi:10.7753/IJCATR1012.1014.

22. Demetriou D, Lockhat Z, Brzozowski L, Saini KS, Dlamini Z, Hull R. The convergence of radiology and genomics: Advancing breast cancer diagnosis with radiogenomics. Cancers. 2024 Mar 6;16(5):1076.

23. Mayegun KO. Multilayered analytics models for dynamic risk assessment in global financial accounting and audit systems. *International Journal of Research Publication and Reviews*. 2025 Jun;6(6):829-849. doi:10.55248/gengpi.6.0625.2025.

24. Waqas A, Tripathi A, Ramachandran RP, Stewart PA, Rasool G. Multimodal data integration for oncology in the era of deep neural networks: a review. Frontiers in Artificial Intelligence. 2024 Jul 25;7:1408843.

25. Otoko J. Economic impact of cleanroom investments: strengthening U.S. advanced manufacturing, job growth, and technological leadership in global markets. Int J Res Publ Rev. 2025;6(2):1289-1304. doi: https://doi.org/10.55248/gengpi.6.0225.0750

26. Zhou H, Zhou F, Zhao C, Xu Y, Luo L, Chen H. Multimodal data integration for precision oncology: Challenges and future directions. arXiv preprint arXiv:2406.19611. 2024 Jun 28.

27. Yan J, Sun Q, Tan X, Liang C, Bai H, Duan W, Mu T, Guo Y, Qiu Y, Wang W, Yao Q. Image-based deep learning identifies glioblastoma risk groups with genomic and transcriptomic heterogeneity: a multi-center study. European Radiology. 2023 Feb;33(2):904-14.

28. Atanda ED. Dynamic risk-return interactions between crypto assets and traditional portfolios: testing regime-switching volatility models, contagion, and hedging effectiveness. *International Journal of Computer Applications Technology and Research.* 2016;5(12):797–807.

29. Ibitoye J. Zero-Trust cloud security architectures with AI-orchestrated policy enforcement for U.S. critical sectors. International Journal of Science and Engineering Applications. 2023 Dec;12(12):88-100. doi:10.7753/IJSEA1212.1019.

30. Takuro KO. Analyzing Intellectual Property Rights adaptation to Artificial Intelligence-created works and automated innovation in the global knowledge economy. *International Journal of Computer Applications Technology and Research*. 2021;10(12):414-424. doi:10.7753/IJCATR1012.1014.

31. Shariaty F, Pavlov V, Baranov M. AI-Driven Precision Oncology: Integrating Deep Learning, Radiomics, and Genomic Analysis for Enhanced Lung Cancer Diagnosis and Treatment: F. Shariaty et al. Signal, Image and Video Processing. 2025 Sep;19(9):693.

32. Hao Y, Cheng C, Li J, Li H, Di X, Zeng X, Jin S, Han X, Liu C, Wang Q, Luo B. Multimodal Integration in Health Care: Development With Applications in Disease Management. Journal of medical Internet research. 2025 Aug 21;27:e76557.

33. Daniel ONI. TOURISM INNOVATION IN THE U.S. THRIVES THROUGH GOVERNMENTBACKED HOSPITALITY PROGRAMS EMPHASIZING CULTURAL PRESERVATION, ECONOMIC GROWTH, AND INCLUSIVITY. International Journal Of Engineering Technology Research & Management (IJETRM). 2022Dec21;06(12):132–45.

34. Padmaja C, Ramacharan S, Krishnan SB, Chakrabarti P. Breast Cancer Diagnostics: Integrating Deep Learning and Radiogenomics. In2025 Third International Conference on Networks, Multimedia and Information Technology (NMITCON) 2025 Aug 1 (pp. 1-5). IEEE.

35. Otoko J. Microelectronics cleanroom design: precision fabrication for semiconductor innovation, AI, and national security in the U.S. tech sector. Int Res J Mod Eng Technol Sci. 2025;7(2)

36. Haq IU, Mhamed M, Al-Harbi M, Osman H, Hamd ZY, Liu Z. Advancements in Medical Radiology Through Multimodal Machine Learning: A Comprehensive Overview. Bioengineering. 2025 Apr 30;12(5):477.

37. Takuro KO. Exploring cybersecurity law evolution in safeguarding critical infrastructure against ransomware, state-sponsored attacks, and emerging quantum threats. *International Journal of Science and Research Archive*. 2023;10(02):1518-1535. doi:10.30574/ijsra.2023.10.2.1019.

38. Sompura DU, Tripathy BK. Deep Learning-Based Techniques for Detection of Cancer. InArtificial Intelligence in Human Health and Diseases 2025 Jul 18 (pp. 355-377). Singapore: Springer Nature Singapore.

39. Rumbidzai Derera. HOW FORENSIC ACCOUNTING TECHNIQUES CAN DETECT EARNINGS MANIPULATION TO PREVENT MISPRICED CREDIT DEFAULT SWAPS AND BOND UNDERWRITING FAILURES. International Journal of Engineering Technology Research & Management (IJETRM). 2017Dec21;01(12):112–27.

40. Hussain D, Al-Masni MA, Aslam M, Sadeghi-Niaraki A, Hussain J, Gu YH, Naqvi RA. Revolutionizing tumor detection and classification in multimodality imaging based on deep learning approaches: Methods, applications and limitations. Journal of X-Ray Science and Technology. 2024 Jul;32(4):857-911.

41. Mayegun KO. Architecting AI-augmented sovereign risk models by integrating climate-energy stressors, macroeconomic indicators, and cross-border cybersecurity intelligence frameworks. *International Journal of Computer Applications Technology and Research*. 2023;12(12):234-251. doi:10.7753/IJCATR1212.1023.

42. Okunlola FO, Adetuyi TG, Olajide PA, Okunlola AR, Adetuyi BO, Adeyemo-Eleyode VO, Akomolafe AA, Yunana N, Baba F, Nwachukwu KC, Oyewole OA. Biomedical image characterization and radio genomics using machine learning techniques. InMining Biomedical Text, Images and Visual Features for Information Retrieval 2025 Jan 1 (pp. 397-421). Academic Press.

43. Emmanuel Damilola Atanda. EXAMINING HOW ILLIQUIDITY PREMIUM IN PRIVATE CREDIT COMPENSATES ABSENCE OF MARK-TO-MARKET OPPORTUNITIES UNDER NEUTRAL INTEREST RATE ENVIRONMENTS. International Journal Of Engineering Technology Research & Management (IJETRM). 2018Dec21;02(12):151–64.

44. Ibitoye JS. Securing smart grid and critical infrastructure through AI-enhanced cloud networking. International Journal of Computer Applications Technology and Research. 2018;7(12):517-529. doi:10.7753/IJCATR0712.1012.

45. Mayegun KO. Advancing secure federated machine learning for multinational defense finance consortia using encrypted AI-driven geospatial and sensor data. *International Journal of Science and Engineering Applications*. 2024;13(12):39-54. doi:10.7753/IJSEA1312.1010.

46. Muneer A, Waqas M, Saad MB, Showkatian E, Bandyopadhyay R, Xu H, Li W, Chang JY, Liao Z, Haymaker C, Soto LS. From Classical Machine Learning to Emerging Foundation Models: Review on Multimodal Data Integration for Cancer Research. arXiv preprint arXiv:2507.09028. 2025 Jul 11.

47. Chen Y, Chen D, Liu X, Jiang H, Wang X. Deep Learning-Driven Multimodal Integration of miRNA and Radiomic for Lung Cancer Diagnosis. Biosensors. 2025 Sep 16;15(9):610.

48. Ibitoye J, Fatanmi E. Self-healing networks using AI-driven root cause analysis for cyber recovery. International Journal of Engineering and Technical Research. 2022 Dec;6: [pages unavailable]. doi:10.5281/zenodo.16793124.

49. Schneider L, Laiouar-Pedari S, Kuntz S, Krieghoff-Henning E, Hekler A, Kather JN, Gaiser T, Froehling S, Brinker TJ. Integration of deep learning-based image analysis and genomic data in cancer pathology: A systematic review. European journal of cancer. 2022 Jan 1;160:80-91.

50. Menegotto AB, Cazella SC. Multimodal deep learning for computer-aided detection and diagnosis of cancer: theory and applications. InEnhanced Telemedicine and e-Health: Advanced IoT Enabled Soft Computing Framework 2021 May 10 (pp. 267-287). Cham: Springer International Publishing.